\documentclass[letterpaper, 10 pt, conference]{ieeeconf}  % Comment this line out if you need a4paper

\IEEEoverridecommandlockouts                              % This command is only needed if 
\usepackage{times}
\pdfoutput=1 % For ARXIV to ensure that we use pdflatex

\usepackage{multicol}
\usepackage{multirow}

\usepackage[utf8]{inputenc} % allow utf-8 input
\usepackage[T1]{fontenc}    % use 8-bit T1 fonts
\usepackage{url}            % simple URL typesetting
\usepackage{booktabs}       % professional-quality tables
\usepackage{amsfonts}       % blackboard math symbols
\usepackage{nicefrac}       % compact symbols for 1/2, etc.
\usepackage{microtype}      % microtypography

\usepackage{wrapfig}
\usepackage{caption}

\newcommand{\senet}[0]{\textbf{SE3-Net}}
\newcommand{\senets}[0]{\textbf{SE3-Nets}}
\renewcommand{\senet}[0]{\textsc{SE3-Net}}
\renewcommand{\senets}[0]{\textsc{SE3-Nets}}

\newcommand{\sref}[1]{Sec. \ref{#1}}
\newcommand{\figref}[1]{Fig.\ref{#1}}

\usepackage[disable,textsize=tiny]{todonotes}
\usepackage[caption=false]{subfig}

\usepackage{amsmath}
\usepackage{mathtools}
\usepackage{cancel}
\usepackage{amssymb}
\usepackage{color}
\usepackage{makecell}

\DeclareMathOperator*{\SE3}{\mathbf{SE}(3)}

\usepackage[bookmarks=true]{hyperref}
\hypersetup{
    colorlinks=true,
    linkcolor=blue,
    filecolor=magenta,      
    urlcolor=cyan,
}
\usepackage{cite}

\title{SE3-Nets: Learning Rigid Body Motion using Deep Neural Networks}

\author{Arunkumar Byravan and Dieter Fox \\% <-this % stops a space
Department of Computer Science \& Engineering \\
University of Washington, Seattle
}

\begin{document}

\maketitle
\thispagestyle{empty}
\pagestyle{empty}

\begin{abstract}
  We introduce \senets\, which are deep neural networks designed to model and learn rigid body
  motion from raw point cloud data. Based only on sequences of depth images
  along with action vectors and point wise data associations, \senets\ learn to
  segment effected object parts and predict their motion resulting from the
  applied force.  Rather than learning point wise flow vectors, \senets\ predict
  $\SE3$ transformations for different parts of the scene. Using simulated depth
  data of a table top scene and a robot manipulator, we show that the structure
  underlying \senets\ enables them to generate a far more consistent prediction
  of object motion than traditional flow based networks.  Additional experiments
  with a depth camera observing a Baxter robot pushing objects on a table show
  that \senets\ also work well on real data.
\end{abstract}

\section{Introduction}

The ability to predict how an environment changes based on forces applied to it is fundamental for a robot to achieve specific goals. For instance, in order to arrange objects on a table into a desired configuration, a robot has to be able to reason about where and how to push individual objects, which requires some understanding of physical quantities such as object boundaries, mass, surface friction, and their relationship to forces.
A standard approach in robot control is to use a physical model of the environment and perform optimal control to find a policy that leads to the goal state.  For instance, extensive work utilizing the MuJoCo physics engine \cite{todorov2012} has shown how strong physics models can enable solutions to control problems in complex and contact-rich environments \cite{geoffroy2014}.  A shortcoming of such models is, however, that they rely on very accurate estimates of the state of the system~\cite{zhouconvex}. Unfortunately, estimating values such as the mass distribution and surface friction of an object using visual information and force feedback is extremely difficult. This is one of the main reasons why humans are still far better than robots at performing even simple tasks such as pushing an arbitrary object along a desired trajectory. Humans achieve this even though their control policies are informed only by approximate, intuitive notions of physics~\cite{battaglia2013simulation,sanborn2013reconciling}. Research has shown that humans learn these mental models from a young age, potentially by observing the effect of actions on the physical world~\cite{baillargeon2004infants}.

In this work, we explore the use of deep learning to model this concept of "physical intuition", learning a model that predicts changes to the environment based on specific actions.  Here, we focus on modeling the motion of systems of rigid bodies, such as predicting how an object on a table moves when being pushed by a robot manipulator.  Importantly, we want to learn predictive models from sequences of \emph{raw 3D point clouds} observed with a depth camera along with continuous action vectors (such as the velocities or torques applied to a robot's joints). Supervision for learning is only provided via point-wise associations between consecutive point clouds, no higher level information such as object segmentation is provided to the learner.  While a standard deep network architecture can be trained on such data to predict the 3D motion of individual observed points, such a vanilla network architecture is not able to learn and represent an explicit notion of \emph{objects and their motion}, which is very useful for control tasks and for higher-level reasoning. 

To overcome this limitation, we introduce \senets, which learn to segment a scene into "salient" objects and predict the motion of these objects under the effect of applied actions. \senets\ represent motion in the environment as a set of $\SE3$ transforms\footnote{$\SE3$ refers to the Special Euclidean Group representing 3D rotations and translations: $\{ R, t | R \in \mathbf{SO}(3), t \in \mathbf{R}^3 \}$},
which are widely used in robotics, computer vision, and graphics to model rigid body motion.  Key to \senets\ is the notion of disentangling the motion of the objects (the \textbf{What?}) from their location in the environment (the \textbf{Where?}).  \senets\ do this by explicitly predicting a set of "$k$" $\SE3$ transforms to encode the motion, and "$k$" dense pointwise masks that specify the contribution of each $\SE3$ towards a 3D point. Finally, the network combines the $\SE3$s, their masks, and the 3D input through a differentiable \emph{transform} layer that blends the motions to produce a predicted output point cloud.

In the absence of any constraints, \senets\ can represent an arbitrary per-point 3D motion. To bias the network to model rigid motion, we adapt a weight sharpening technique used in \cite{whitney2016disentangled} and show that this results in a segmentation of the environment into distinct objects along with their rigid motion. We show results on three simulated scenarios where our system predicts the motion of a varying number of rigid objects under the effect of applied forces, and a robot arm with four actuated joints. We present experiments testing different parts of our network and show results highlighting the robustness of \senets\ to different types of noise, similar to those found in real world data.  We also show the capability of \senets\ to operate on real world data collected using the Baxter robot pushing objects on a table.

The main contributions of this paper are as follows. We introduce \senets, a deep neural network architecture that models scene dynamics by segmenting the scene into distinct objects and jointly predicting their rigid body ($\SE3$) motion. We show that \senets\ can learn to do this solely based on sequences of control, raw depth camera data and point-wise associations, without the need for explicit segmentation information in the training data. We also provide substantial experimental evidence that \senets\ outperform standard deep learning baselines, and can be applied to real robot data.

This paper is organized as follows.  After discussing related work, we introduce \senets\ in Section~\ref{sec:approach}, followed by experimental evaluation and discussion.

%%% Local Variables:
%%% mode: latex
%%% TeX-master: "paper"
%%% End:

\section{Related work}

\textbf{Robotics:} As mentioned before, many optimal control techniques require a "dynamics" model that predicts the effect of actions on the state of the system~\cite{toussaint2009robot}. Early work on learning dynamics models from data focused on low-dimensional state and control representations \cite{deisenroth2011pilco}. More recent work has looked at learning forward rigid-body dynamics models assuming tracking information is available \cite{kopicki2016learning}. In contrast to these methods, Boots et al.~\cite{boots2014learning} learn a model using Predictive State Representations to predict raw depth images given a history of prior images and control. There is also a large body of work investigating how robots can interact with objects for manipulation or object exploration tasks, as summarized in a recent survey~\cite{Boh16Int}. The focus of this area, however, is not on learning predictive motion models from raw data, as we present in this paper. Recently, deep models have been used for learning dynamics models in robotics and reinforcement learning, by mapping raw pixel images to low-dimensional (latent) encodings on top of which standard optimal control methods are applied~\cite{watter2015embed, wahlstrom2015pixels}. In a similar flavor, \senets\ explicitly model the dynamics of the scene as rigid body motion of salient objects, jointly learning object segmentation and motion prediction.

%\textbf{Deep learning in robotics:} Deep learning models have recently been very successful on a wide variety of tasks in computer vision such as classification, semantic labeling and object recognition. Deep models have also been used for learning dynamics models in robotics and reinforcement learning, by mapping raw pixel images to low-dimensional (latent) encodings on top of which standard optimal control methods are applied~\cite{watter2015embed, wahlstrom2015pixels}. \senets\ can also be used similarly and have the added advantage of operating in the interpretable 3D physical world. 

\textbf{Physics prediction:} Recent work in deep learning has looked at the problem of predicting the stability of stacked blocks \cite{lerer2016learning}, ball motion \cite{fragkiadaki2015learning}, and object dynamics in images \cite{mottaghi2016newtonian}. In particular, work by Agrawal et al.~\cite{agrawal2015learning} and Finn et al.~\cite{finn2016unsupervised} are closely related, both focusing on modeling the effect of robot actions on a scene, albeit using RGB images. Unlike \cite{agrawal2015learning}, we explicitly predict a dense point cloud through our forward model, encoding the motion using $\SE3$ transforms and masks. Similar to \cite{finn2016unsupervised}, we composite motion using predicted masks with two main differences: we use $\SE3$ transforms in 3D to capture complex out of plane motions and our masks are sharpened to enforce the rigid-body assumption while improving prediction accuracy.

%These methods mostly predict low-dimensional outputs like ball velocity \cite{fragkiadaki2015learning} or the probability of falling \cite{li2016fall}. One exception is the work by Lerer et al.~\cite{lerer2016learning}, which predicts images of a tower of blocks, but their network operates on RGB images and has no specific notion of an action or forces. 

\textbf{Predicting 3D rotation:} Related work in the computer vision literature has looked at the problem of predicting 3D motion, primarily rotation between pairs of images \cite{kulkarni2015deep, yang2015weakly}. Differing from most of this work, we operate on 3D data, incorporate continuous actions, and explicitly predict rigid body motion and object masks. Independently in parallel to our work, Handa et al.~\cite{handa2016gvnn} proposed deep models using rigid body transforms for depth image registration and alignment, but do not model object motion or the effect of actions.

%Perhaps, the most similar work to ours is by Yang et al.~\cite{yang2015weakly}, who proposed a deep recurrent model that predicts the change in an encoded pose vector through the effect of a one-hot action vector to render rotated color images of objects. Differing from their work, we operate on 3D data, use a continuous action vector, and explicitly predict rigid body motion and object masks. 

\textbf{Attentional models and disentangling representations:} A related line of work to ours is the idea of attentional mechanisms \cite{gregor2015draw, ba2014multiple} which focus on parts of the environment related to task performance and the concept of disentangling representations \cite{kulkarni2015deep, yang2015weakly}, which aim to separate variations in the environment. Our model has a differentiable, dense, point-wise attender that learns to focus on parts of the environment where motion occurs, using these cues to segment objects. Also central to our model is the idea of disentangling the motion of the object from its location. Finally, our model is related to the Spatial Transformer network \cite{jaderberg2015spatial}, though we model the effect of actions, use dense attention and restrict ourselves to $\SE3$ transformations.

%%% Local Variables:
%%% mode: latex
%%% TeX-master: "paper"
%%% End:

\section{\senets}
\label{sec:approach}
Given a 3D point cloud ($X$) from a depth sensor and an $n$-dimensional continuous action $u$ as input, \senets\ model the scene dynamics as rigid body motions of the constituent objects to generate a transformed point cloud $Y$:
\begin{equation}
Y = f(X, u) \ | \  f:= \{R_i, t_i, M_i\}, i = 1 \dots k
\end{equation}
In essence, \senets\ decompose the scene into $k$ rigid objects, predicting per object a mask $M$ that attends to parts of the scene containing the object and a rigid body transform $[R,t] \in \SE3$ that quantifies the object's motion. Note that in our setting, $k$ is a pre-specified network parameter that limits the number of distinctly moving objects or parts (including background which has no motion).

\figref{fig:network} shows the general architecture of \senets\ . There are three major components: an encoder that generates a joint latent state given the input point cloud $X$ and the control $u$, a decoder that predicts the object masks with the corresponding transforms and a final transformation layer that generates the transformed point cloud $Y$. 

\subsection{Encoder}
The encoder has two parts: a convolutional encoder which generates a latent state from the point cloud $X$ (represented as a 3-channel image) and a fully connected network that encodes the control vector $u$. We adopt a late fusion architecture and concatenate the outputs of these two parts to produce the final encoding, which is used by the decoder for further predictions. 

%where $[R,t] \in \SE3$ is a rigid body transform, $M$ represents a notion of objects and $k$ is a pre-specified network parameter which limits the number of distinctly moving objects or parts (including background which has no motion).

%\figref{fig:network} shows the general architecture of \senets\. Our network takes a 3D point cloud ($X$) shaped as a 3-channel image and an $n$-dimensional continuous vector ($u$) as input and generates a transformed 3D point cloud ($Y$) as output. There are three major components to our network: an encoder, a decoder and a final transform layer. The encoder is a straightforward late-fusion convolutional/fully-connected architecture that processes the input point cloud and controls separately to produce low-dimensional encoded vectors. We concatenate the encoded vectors to produce a single joint encoding which is used by the rest of the network.

\subsection{Decoder}
The decoder decomposes the motion prediction problem into two sub-problems: identifying and grouping together points that move together (we call this grouping an "object" or a "motion-class") and subsequently predicting the $\SE3$ transformation parameters that quantify the object's motion. 

%The decoder uses the encoder output to predict motion in the scene by separating into two parts: the mask decoder, which predicts each object's location in the scene through the mask $M$, and the $\SE3$ decoder which predicts the rigid transformation parameters for each object.

%The concatenated encoding vector is used by the decoder to predict the motion in the scene by separating it into two parts: the mask, which attends to where motion occurs, and the $\SE3$ transformation parameters, which specify the motion itself.  

\begin{figure*}
  \begin{center}
    \centering
    \vspace{1mm}
    \includegraphics[width=0.8\textwidth]{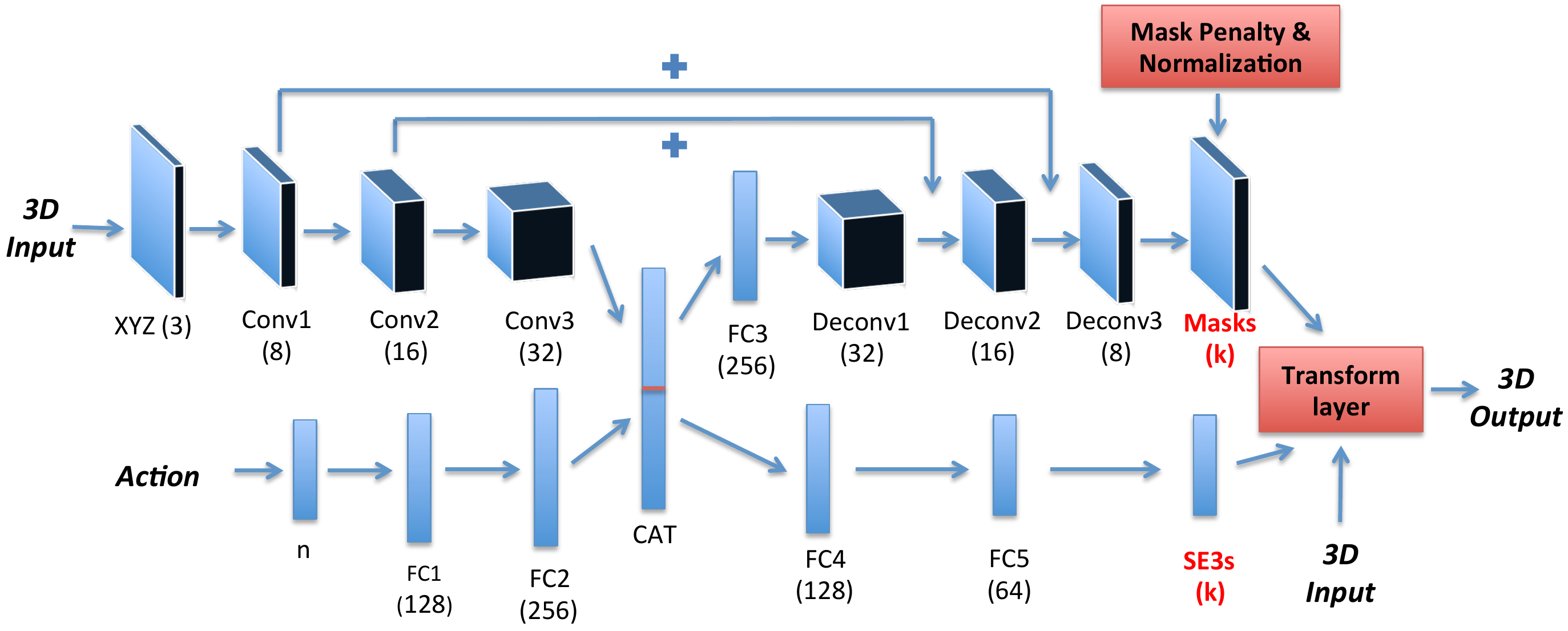}
    	\vspace{-1mm}
    	\vspace*{-1ex}
    \caption{\small{\senet\ architecture. Input is a 3D point cloud and an $n$-dimensional action vector (bold-italics), both of which are encoded and concatenated to a joint feature vector (CAT). The decoder uses this encoding to predict "$k$" object masks $M$ and "$k$" $\SE3$ transforms which are used to transform the input cloud via the "Transform layer" to generate the output. Mask weights are sharpened and normalized before use for prediction. Conv = Convolution, FC = Fully Connected, Deconv = Deconvolution, CAT = Concatenation }}\vspace*{-5ex}
    \label{fig:network}
  \end{center}
\end{figure*} 

\textbf{Predicting motion masks: }
The mask decoder attends to parts of the scene that exhibit motion, grouping points that move together to form objects. As an example, all points belonging to a rigid object can be grouped together as they move with it. Presupposing that the scene has $k$ distinctly moving objects, we can formulate this as a $k$-class labeling problem where each input point can belong to one of the $k$ motion-classes. Unfortunately, this formulation is non-differentiable due to the discreteness of the labeling. Instead, we relax it to allow each point to belong to multiple motion classes, quantified by a per-point probability distribution $M^j$ over the $k$ motion-classes:
 \vspace{-2mm}
\begin{equation}
M^j = \{m_{1j}, m_{2j},\dots, m_{kj}\} \ | \sum_{i=1}^{k} m_{ij} = 1; 
\label{eq:maskprob} 
 \vspace{-1mm}
\end{equation}
allowing each point $j$ to smoothly interpolate between multiple motions. 

We use a de-convolutional architecture to compute the dense object masks, generating $k$ masks (represented as a $k$-channel image) at the input resolution. Following recent work \cite{long2015fully}, we use a skip-add architecture wherein we add the convolutional layer outputs to the de-convolutional layer inputs. This gives us sharper reconstructions of object shapes and contours, improving overall performance.

\textbf{Predicting rigid transforms: }
As mentioned before, we represent motion in the environment using 3D rigid body transforms. A 3D rigid body transform $[R,t] \in \SE3$ can be specified by a rotation $R \in \mathbf{SO(3)}$ and a translation $t \in \mathbf{R}^3$. A 3D point $x$ affected by this transformation moves to:
%\vspace{-1mm}
\begin{align}
x^\prime &= Rx + t
\vspace{-4mm}
\end{align}
We represent rotations using a 3-parameter axis-angle transform $a \in \mathbf{R}^3$, with $||a||_2 = \theta$, the magnitude of rotation. 

The $\SE3$ decoder predicts $k$ $\SE3$ transforms, one for each of the $k$ motion-classes (including background). We use a fully connected network to predict these transforms.

\subsection{Transform layer}
Given the predicted $\SE3$ transforms and masks, the transform layer produces a blended output point cloud from the input points:
 \vspace{-5mm}
\begin{equation}
y_j = \sum_{i=1}^{k} m_{ij} \left( R_i  x_j + t_i \right)  
 \vspace{-1mm}
\label{eq:finaltfm} 
\end{equation}
where $y_j$ is the 3D output point corresponding to input point $x_j$. Eqn.~\ref{eq:finaltfm} computes a convex combination of the transformed input points, transformed by each of the $k$ $\SE3$ transforms with weights given by the object mask. As a consequence of our relaxation for the mask (Eqn.~\ref{eq:maskprob}), the effective transform on a given point is generally not in $\SE3$ as Eqn.~\ref{eq:finaltfm} blends in 3D space rather than in the space of $\SE3$ transforms. On the other hand, we now have the flexibility to represent both rigid and non-rigid motions through a combination of the transforms and the oject masks. Additionally, we avoid the potential singularities that can arise from blending in $\SE3$ space. In spite of the advantages, using the current framework to model rigid motion without any explicit regularization can lead to over-fitting and blurry predictions. To avoid this, we encourage the network to predict rigid motions through a form of regularization on the object mask.

\textbf{Enforcing Rigidity:}
A simple way to restrict the network to predict rigid motions is to force the per-point mask probability vector $M^j$ to make a binary decision over the $k$ predicted transforms instead of blending. As mentioned before, a naive formulation can lead to non-differentiability. Instead, we smoothly encourage the mask weights towards a binary decision using a form of weight sharpening \cite{whitney2016disentangled}:
%
%\vspace{-1mm}
%\begin{equation}
%{m_{ij}}^\prime = \dfrac{( m_{ij} + \epsilon )^\gamma}{\sum_k {m_{kj}^\gamma}}; \ \ \epsilon \sim \mathcal{N}(0, \sigma^2)
% \vspace{-1mm}
%\end{equation}
%
%\vspace{-1.5ex}
\begin{align}
\forall i = 1 \dots k, \ {m_{ij}}^\prime &= ( m_{ij} + \epsilon )^\gamma; \ \ \epsilon \sim \mathcal{N}(0, \sigma^2) \nonumber \\
{m_{ij}}^{\prime\prime} &=  \dfrac{{m_{ij}}^\prime}{\sum_k {m_{kj}}^\prime}
\vspace{-2ex}
\end{align}
where the noise $\epsilon$ is sampled from a Gaussian, $\gamma, \sigma$ are proportional to the training epoch. In practice, the combination of the noise and growing exponent forces the network to push its decisions apart, resulting in nearly binary distributions at the end of training. Finally, at test time, the network segments the input point cloud $X$ into $k$ distinct objects, predicts their motion and applies a rigid transform to each input point (Eqn.~\ref{eq:finaltfm}), to generate the transformed output point cloud $Y$.

%%% Local Variables:
%%% mode: latex
%%% TeX-master: "paper"
%%% End:

\section{Evaluation}
We evaluate \senets\ on how well they predict motion on multiple simulated tasks using the Gazebo physics simulator and real world data collected using a Baxter robot poking at objects on a table. We first present results on simulated data, followed by tests on the robustness of the network to different types of noise and hyper-parameter choices and finally discuss results on real world data. 

%The task of the network is to predict the motion of a robot manipulator and rigid objects based on control commands and a moving ball, respectively.  The object motion scenarios require the network to learn to segment effected objects from the scene and predict their motion, which depends on which location on the object is being hit by the ball.  

%The other scenarios test how well the network is able to learn the articulation of a robot manipulator resulting from input controls.  All our datasets contain sequences of simulated point clouds taken from a depth camera looking at the environment from a fixed viewpoint (for each task) along with the state of all the objects and dense 3D optical flow between point cloud pairs. In all our tasks, the network predicts a target 5 timesteps into the future (0.15 sec) given the current point cloud (3-channel 240 x 320 image) and control. We assume that the control is held fixed for this duration.

\subsection{Simulated data collection}
We set up four simulated tasks using the Gazebo physics simulator, consisting of scenes where a fixed camera looks at rigid bodies moving under the effect of applied forces. In all our settings, the network takes a 3D point cloud (240$\times$320 resolution), and an $n$ dimensional continuous control vector as input and predicts the resulting point cloud 0.15 seconds in the future. We assume that the control is held fixed for this duration. We detail the data collection next.

\textbf{Single Box:} This dataset has 9000 random scenes where a ball collides with a box placed at a random position on a table. In each scene, we place the ball at a random location in front of the box and continuously apply a randomly chosen constant force to the ball, directing it to collide with the box. For a given scene, we run the simulation for one second, record data and discard frames where the box falls off the table. Across scenes, we vary the start pose of the objects and the applied force while keeping the objects' size and mass constant. We also vary the table size to introduce background variations. Across all scenes, we have a total of \textasciitilde 170,000 examples (each example consists of an input control and input \& target point clouds). The control vector $u$ is 10-dimensional, consisting of the ball pose (position \& quaternion) and the applied force ($n$ = 10). We add the ball's orientation to the control to model cases where the ball undergoes spin. 

\textbf{Multiple Boxes:} To test the generalization of the system to different object sizes, masses and number of objects, we generated a second dataset that varies all three at random. Each scene has anywhere from 1-3 objects of varied sizes and proportional mass, where the ball collides with a randomly chosen box. We only consider examples where a single box and the ball are in collision and discard those that involve multiple collisions as it is hard for the system to model such motion without any temporal information. This dataset has \textasciitilde 12,000 different scenes, with a total of \textasciitilde 210,000 examples. The controls are represented the same as the Single Box dataset ($n$ = 10).

\textbf{Baxter:} Our third dataset consists of sequences of depth images looking at a Baxter robot being controlled to move its right arm randomly. In each scene, we apply a constant, randomly chosen velocity to 1-4 randomly chosen joints on the robot's right arm. Each scene lasts for one second, after which we bring the arm to a rest. We randomly reset the pose of the arm once every 20 scenes. In total, this dataset has \textasciitilde 11,000 scenes with \textasciitilde 220,000 examples. The controls for this task are the commanded joint velocities, a 14-dimensional vector in which 1-4 values are non-zero ($n$ = 14).

\textbf{Household Objects:} The final simulated dataset tests the generalization of the system to irregular object shapes. We use 11 household objects from the LineMOD dataset \cite{hinterstoisser2012accv}. Each scene has 1-4 objects randomly placed on a table, with the ball colliding against a randomly chosen object. In total, we have 10,000 random scenes with \textasciitilde 200,000 examples. The controls are similar to the box datasets ($n$ = 10). This dataset is particularly challenging as it has significant amounts of toppling and fast rotations due to the objects' irregular shapes.

\subsection{Training}
We implemented our system using the deep learning package Torch \cite{collobert2011torch7}. We trained our networks using the ADAM optimization method \cite{kingma2014adam} along with Google's Batch Normalization technique \cite{ioffe2015batch} to speed up training. At the start of training, we initialize the layer predicting $\SE3$ transforms to predict the identity transform which we found to improve convergence. We initially set the weight sharpening penalty to zero and slowly ramp up the noise parameter $\sigma$ and the exponent $\gamma$ till they reach a preset maximum. The number of objects $k$ is chosen apriori: $k = 5$ for the Baxter dataset (4 joints + background) and $k = 3$ for all other datasets (ball + object + background). 

\textbf{Comparison:} We train three variants of $\senets$ and compare against two baseline networks that predict 3D scene flow as well as a baseline that always predicts zero motion:
\vspace{-0.5ex}
\begin{itemize}
\item \textit{Ours: } \senet\ from Fig.~\ref{fig:network}, with \textasciitilde 1.4 million parameters.
\item \textit{Ours (Large): } Bigger version of the \senet\ with 5-Conv/Deconv layers and \textasciitilde 6x as many parameters. 
\item \textit{No Penalty: } \senet\ without any weight sharpening to enforce binarization of the object masks. 
\item \textit{Flow: } Network trained to predict dense 3D optical flow directly using a Conv/Deconv architecture similar to the original \senet , but without the $\SE3$ prediction module and the transform layer. 
\item \textit{Flow (Large): } Bigger version of the flow network with 5-Conv/Deconv layers and \textasciitilde 8x as many parameters as the original \senet.
\item \textit{No Motion: } Baseline that always predicts zero motion.
\end{itemize}
Both the small networks use strided convolution with no pooling, while the large networks use max pooling instead of striding. All networks use the Parametric-ReLU non-linearity. We trained the networks for 50k iterations on the single box and Baxter datasets and for 75k iterations on the other two datasets with a 70:30 train/test split. Depending on the network size, training takes anywhere between a few hours to half a day on an NVIDIA Titan X GPU.

\textbf{Training targets:} Given an input point cloud $X$ and control $u$, \senets\ predict the output point cloud $Y$ by transforming the input points (Eqn.~\ref{eq:finaltfm}). We would like these predictions to match the ground truth targets $Y^\prime$, which specify the true position of our input points at a future time. In order to compute these targets though, we need to track and associate a given input point $x_j$ across multiple depth images. In this work, we assume that these associations are given to us at training time, either by the physics simulator or an object tracker (in case of real data). We discuss ways to overcome this assumption later in \sref{sec:discussion}. 

\textbf{Evaluation Metric:} We report Mean Squared Error (MSE) between the predicted 3D scene flow (computed as the difference between the input and the predicted point clouds) and ground truth, averaged across points with non-zero ground truth flow. This metric takes into account errors in both the mask and $\SE3$ predictions, and is also the loss function used to train the flow networks. 
\begin{figure*}
  \begin{center}
  	\vspace{1.5mm}
    \centering
    \includegraphics[width=0.9\textwidth]{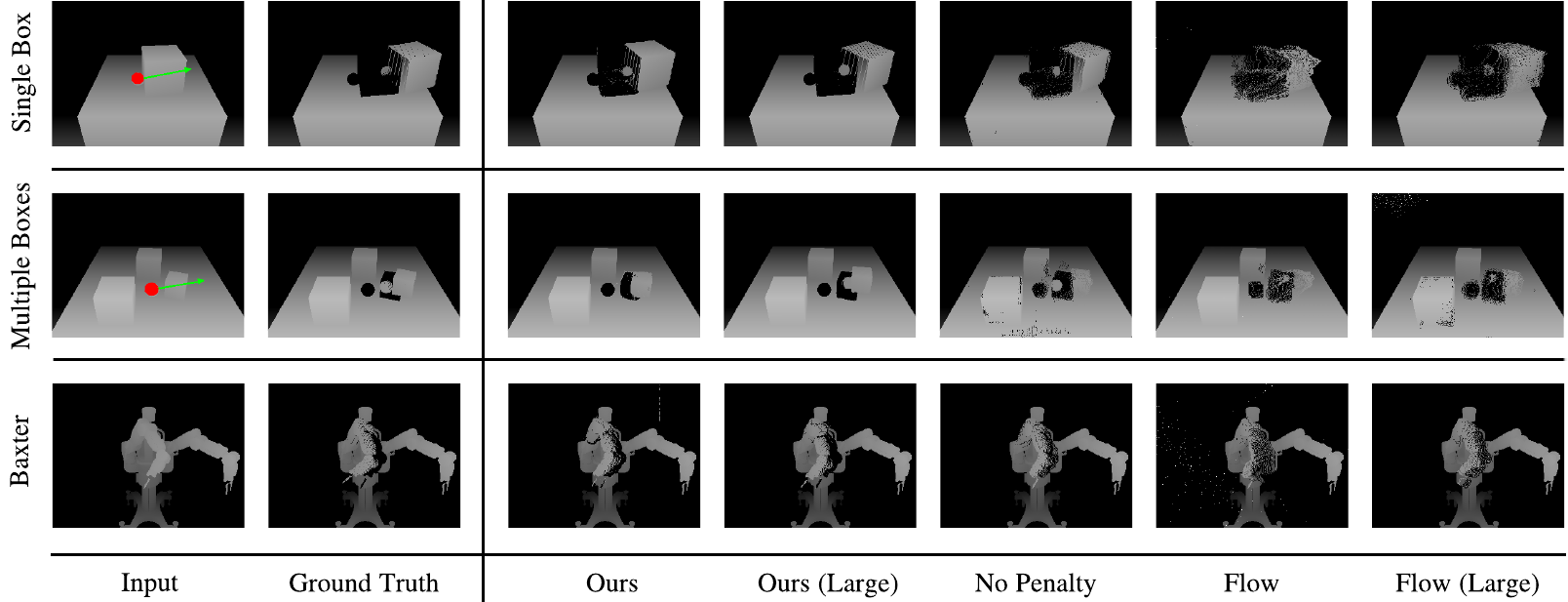}\vspace*{-1ex}
    \caption{\small{Prediction results for three simulated datasets. All images (except first column on the left) were rendered by projecting the predicted 3D point cloud to 2D using the camera parameters and rounded off to the nearest pixel without any interpolation. (From left to right) Input point cloud with the ball highlighted in red and applied force shown in green; ground truth; predictions generated by the different networks. 3D point clouds for the flow networks were computed by adding the predicted flow to the input. The black regions in the images correspond to parts that were occluded in the input and later became visible due to the objects' motion (none of the networks can fill in missing data). Image best viewed in high resolution. For a better understanding of the results, please refer to the supplementary video.}}
    \label{fig:simresults}\vspace*{-2ex}
  \end{center}
\end{figure*} 
\begin{table*}
\centering
\begin{tabular}{| c | c | c | c | c | c | c |}
\hline
Task              		& Ours & Ours (Large) & No Penalty & Flow & Flow (Large) & No Motion \\
\hline
Single Box 		 & 3.73 	& \textbf{1.65 $\pm$ 0.17} & 4.39 & 10.1 & 2.48 $\pm$ 0.22 & 23.24 \\
Multiple Boxes   & 3.22 	& \textbf{1.29 $\pm$ 0.14} & 2.83 &  6.2 & 1.69 $\pm$ 0.17 & 21.84  \\
Baxter           & 0.074 & \textbf{0.057 $\pm$ 0.002} & 0.074 & 0.11  & 0.063 $\pm$ 0.001 & 0.33 \\
 \hline
\end{tabular}
\caption{\small{Average per-point flow MSE (cm) across tasks and networks. Our (Large) network achieves the best flow error compared to baselines even though it is not directly trained to predict flow. "No Motion" results quantify the average magnitude of motion in the datasets (>20 cm for box datasets, < 1 cm for Baxter datasets).}}
\label{tbl:flowerror}\vspace*{-3ex}
\end{table*}
\begin{figure*}
  \begin{center}
    \centering
    \includegraphics[width=0.65\textwidth]{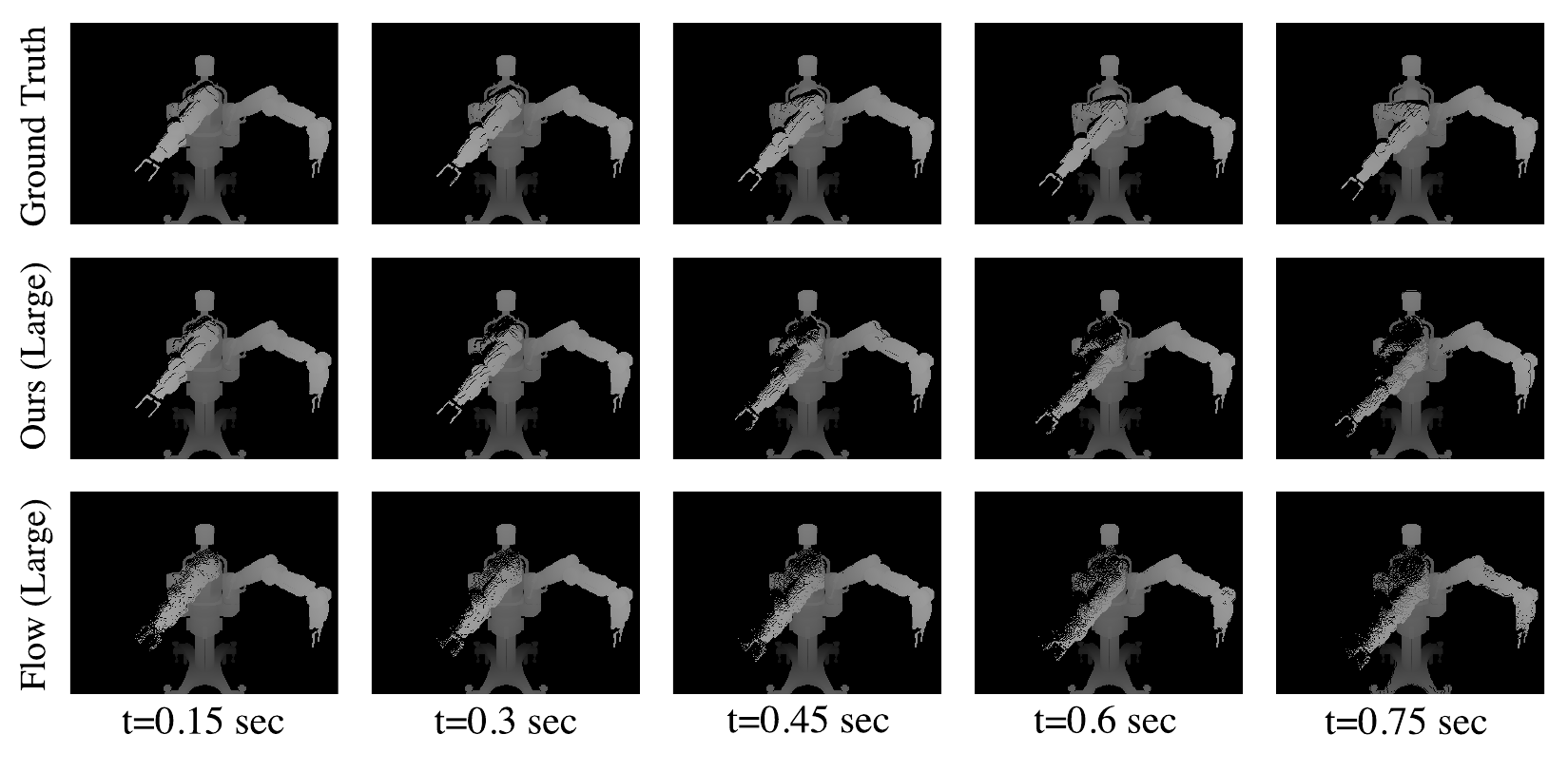}\vspace*{-2ex}
    \caption{\small{Multi-step prediction results obtained by feeding back the output of the network as the input for four consecutive times. Ground truth is reset at the end of each frame.}}
        \label{fig:baxtersequence}\vspace*{-1ex}
  \end{center}
\end{figure*} 
\begin{figure*}
  \begin{center}
    \centering
    \includegraphics[width=0.7\textwidth]{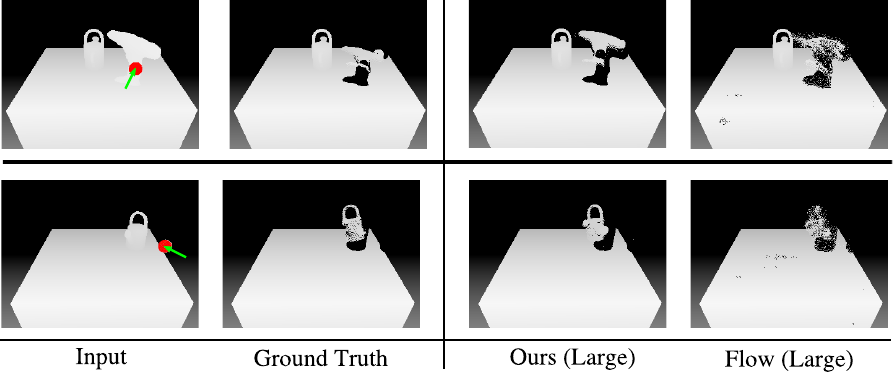}\vspace*{-2ex}
    \caption{\small{Two results from the "Household Objects" dataset, showing topping and sliding motion. Our network segments the objects correctly and predicts consistent motion leading to sharp images while the flow baseline smears the object across the image. Ball highlighted in red.}}
        \label{fig:multiobj}\vspace*{-1ex}
  \end{center}
\end{figure*} 
\begin{table*}
\centering
\vspace{1.5mm}
\begin{tabular}{ | c | c | c | c | c ||  c | c | c | c |}
\hline
 \multirow{3}{*}{Task} & \multicolumn{4}{c||}{Depth Noise (SD = Noise standard deviation)} & \multicolumn{4}{c|}{Data Association Noise} \\
\cline{2-9}
& \multicolumn{2}{ c| }{SD = 0.75 cm} &  \multicolumn{2}{ c|| }{SD = 1.5cm, No depth scaling} & \multicolumn{2}{ c| }{9$\times$9, threshold = $\pm$10cm} &  \multicolumn{2}{ c| }{15$\times$15, threshold = $\pm$20 cm}  \\
\cline{2-9}
              		 & Ours (Large) & Flow (Large) & Ours (Large) & Flow(Large) & Ours (Large) & Flow (Large) & Ours(Large) & Flow (Large)\\
\hline
Single Box 		& \textbf{2.61} 		 & 6.87 					& \textbf{2.70} 		 & 4.31  	& \textbf{1.79} 		& 3.15 		& \textbf{2.80} 		 & 5.32 \\
Multiple Boxes   & \textbf{1.95} 		 & 6.10 					& \textbf{3.56} 		 & 4.42   & \textbf{1.05} 		 & 1.95 		& \textbf{2.48} 		 & 4.26  \\
Baxter           		& 0.073 	 			 & \textbf{0.066}  	& \textbf{0.44} 		 & 0.63 	& \textbf{0.10} 	 	& 0.15  		& \textbf{0.30} 		 & 0.43  \\
  \hline
\end{tabular}\vspace{-1.5mm}
\caption{\small{Average per-point flow MSE (cm) for networks trained with noise added to depth (left four columns) and data associations (right four columns). Our networks' performance degrades gracefully with increasing noise as compared to large errors for the flow baseline.}}
\label{tbl:noise}\vspace*{-5ex}
\end{table*}
\subsection{Results on simulated data}
Table~\ref{tbl:flowerror} reports test results on the first three simulated datasets. Our networks significantly outperform their counterpart small and large flow networks on all tasks. We also did a 5-fold cross validation for the large networks and found that our improvements over the flow baselines are statistically significant. Our networks also achieve a large reduction in prediction error compared to the zero motion baseline (Table~\ref{tbl:flowerror}, last column), even for very large motions (> 20cm per point). In practice, we've seen similar performance on shorter horizons (0.06 sec) with smaller, subtle motions.

\begin{figure*}
  \begin{center}
    \vspace{1ex}
    \centering
    \includegraphics[width=0.6\textwidth]{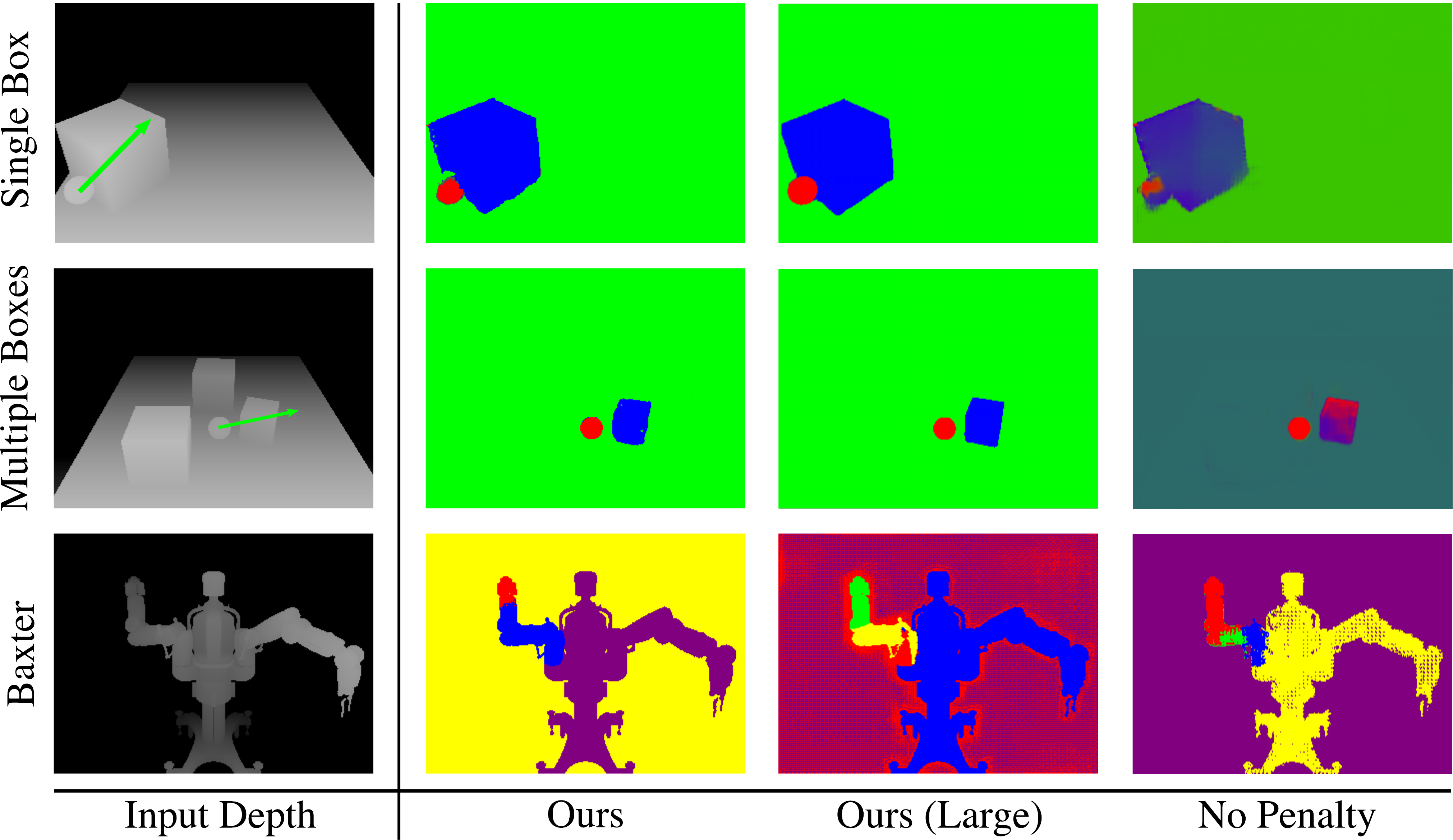}\vspace*{-1ex}
    \caption{\small{Object masks predicted by our networks. Box masks ($k$=3) are rendered directly as RGB images while Baxter masks ($k$=5) are colored based on an arg-max operation across the $k$-mask channels. \senet\ predictions (middle two columns) are near-binary as seen by the distinct coloring for each distinctly moving object (eg: blue for the box, red for the ball). \textbf{No Penalty} masks (right column) have mixed coloring across the scene (eg: indigo for the box), indicating that the masks are not binary. Image best viewed in color.}}
  \label{fig:masks} \vspace{-5.5ex}
  \end{center}
\end{figure*}

Fig.~\ref{fig:simresults} shows some representative predictions made by the networks on three simulated datasets. These results highlight yet another advantage of our approach as compared to a naive flow baseline - the consistency and sharpness of our predictions. By segmenting the scene into distinct objects (Fig.~\ref{fig:masks}) and predicting individual $\SE3$ transforms, our network ensures that points which belong to an object rigidly move together in an interpretable manner. This results in a sharp prediction with very little noise, as compared to the flow networks which do not have any such constraints. With increasing layer depth, the flow networks can somewhat compensate for this, but there is still a significant amount of noise in the predictions resulting in smearing across the canvas (Fig.~\ref{fig:simresults}). Surprisingly, we noticed that the flow networks perform quite poorly on examples where only a few points move such as when just the ball moves in the scene while our networks are able to predict the ball's motion quite accurately. 

We also present mask predictions made by our networks in Fig.~\ref{fig:masks}. The colors indicate that the masks predicted by our networks for the box datasets are near binary (we render the 3-channel masks directly as RGB images). Our network successfully segments the ball and box as distinct objects without any explicit supervision. In practice, we found that it is crucial to give the network examples where the ball moves independently as this provides implicit knowledge that the ball and the box are distinct objects. In cases where the training examples always have the ball in contact with the box, the network had a hard time separating the objects, often masking them out together. For the Baxter dataset, depending on the motion, our network usually segments the arm into 2-3 distinct parts (Fig.~\ref{fig:masks}), often with a split at the elbow. 

In comparison, the "No Penalty" \senet\ rarely predicts binary masks, often blending across different $\SE3$s (rightmost column, Fig.~\ref{fig:masks}). While this leads to some overfit on datasets with large motion such as the Single Box dataset (4.4 cm MSE compared to 3.7 cm for the small \senet\~), it still performs quite well on the other datasets. Interestingly, the "No Penalty" networks significantly outperform the smaller flow networks, hinting that our mask/$\SE3$ decomposition structure (even without the rigidity penalty) better models the underlying problem structure for dense motion prediction.

%We believe that this is because blending allows the network to capture the serial dependence in the Baxter's kinematic chain better. This hints at an interesting approach for modeling non-rigid motions and kinematic chains using the $\SE3$ net, which we discuss further in Sec. \ref{sec:discussion}.
%
Fig.~\ref{fig:multiobj} shows two representative results from testing on the household objects dataset where the network has to deal with complex shaped objects, some with holes. As it is clear, the network can model the dynamics of these objects well, with the resulting predictions being significantly sharper compared to the large flow network, which does very poorly. We have also seen that our network is able to gracefully handle cases where objects topple or undergo large motions. 

Finally, we test the consistency of our network in modeling sequential data by allowing the network to predict scene dynamics multiple steps in the future.  Fig.~\ref{fig:baxtersequence} shows these results for a Baxter sequence where we feed the network's predictions back in as input and fix the control vector for 5 steps into the future. We compare against ground truth and see that our predictions remain consistent across time without significant noise addition. In comparison, the predictions from the large flow network degrade over time as the noise cascades. To get a better understanding of our results, we encourage the readers to look at the supplementary video where we show prediction results for many sequences.

\vspace{-1ex}
\subsection{Robustness}
We perform three additional experiments to test the robustness of our networks to hyper-parameter choices and noisy data. 
\textbf{Sensitivity to number of objects:} In all prior experiments, we have chosen the number of predicted $\SE3$s ($k$) apriori with our knowledge of the datasets. To test the sensitivity of our algorithm to this parameter, we trained our networks setting $k$ to a large number ($k = 8$ for the Baxter dataset and $k = 6$ for the rest). In most cases, we found that the network automatically segments the scene into the correct number of objects, with the remaining mask channels assigned to identity. We also saw little to no performance drop in these experiments. 
%\textbf{Sensitivity to prediction horizon:} To test the robustness of our network to the prediction horizon (0.15 seconds so far), we tested our networks on two smaller horizons (0.06 \& 0.03 seconds). We saw similar performance for both these cases, with slightly poor results on the smallest horizon (0.03 seconds) due to difficulty in disambiguating distinct objects from small motions. 
%\textbf{Sensitivity to initialization:} As mentioned before, we initialize the $\SE3$ decoder to predict identity at the start of training. When relaxing this by doing a random initialization, we saw only a small drop in performance, with an increase in the convergence time.
\textbf{Robustness to depth noise:} To test whether our network is capable of handling the types of noise seen in real depth sensors, we trained networks under two types of depth noise: First, we added gaussian noise with a standard deviation (SD) of 0.75 cm, and scaled the noise by the depth (farther points get more noise) as is common in commodity depth sensors. Second, we increased the noise SD to 1.5cm without scaling by the depth. Table~\ref{tbl:noise} shows the performance of the two large networks under both types of noise: while our performance degrades, we significantly outperform the baseline flow network. Additionally, our network is still able to segment the objects properly in most of our tests.
\textbf{Robustness to noise in data association:} We test how well our networks respond to uncertainty in data association by allowing spurious ground-truth associations when computing our training targets. We allow each point to be randomly associated to any other point in a $m \times m$ window around it, as long as their depth differences are no larger than a threshold. We train in two increasingly noisy settings: associating in a $9 \times 9$ window with a threshold of $\pm$10 cm and in a $15 \times 15$ window with a threshold of $\pm$ 20 cm. Table~\ref{tbl:noise} shows the results of these tests. Our network strongly outperforms the flow baseline, with errors almost half of the flow baseline. While this test does not simulate a systematic association bias, it still shows that our network is robust to uncertain associations.
We believe that the strong structural constraints inherent in our network allow it to average over noise, leading to robust predictions even in highly noisy settings. 

\begin{figure*}
  \begin{center}
    \centering
    	\vspace{1ex}
    \includegraphics[width=0.825\textwidth]{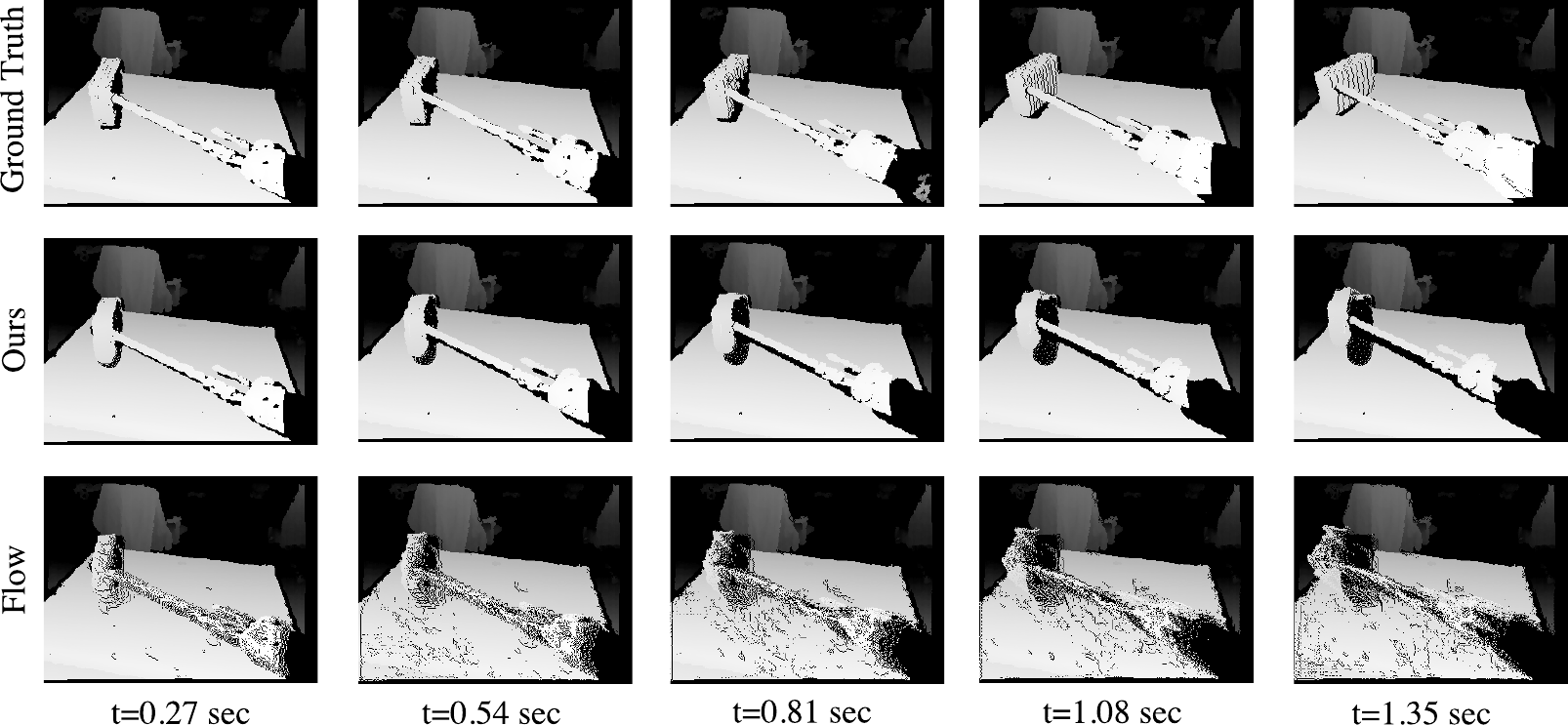}\vspace*{-0.75ex}
    \caption{\small{Multi-step prediction results on real world data collected by poking objects with the Baxter robot. Our network predictions are sharper than the flow baselines highlighting the consistency of the learned segmentation and transforms. }}
        \label{fig:realpokesequence}\vspace*{-6ex}
  \end{center}
\end{figure*} 

\vspace{-1ex}
\subsection{Results on real data}
Finally, we present results from a preliminary evaluation of \senets\ on real world data obtained with  the Baxter robot interacting with three objects in a tabletop scene (a cheezeit box, mustard bottle and pringles can), poking them with a stick attached to it's end-effector. At the start of each poking action, we randomly place a single object on the table. Similar to \cite{agrawal2016learning}, we choose a random direction to poke in, while keeping the end-effector level and at a constant orientation. Each poking action lasts around 5-7 seconds during which we record point clouds from a depth camera mounted on the torso of the robot along with joint encoder data. As a preliminary effort, we collected 67 poke actions with mostly sliding and rotational object motion for a total of 7700 examples. We use the DART motion tracker \cite{schmidt2014dart} to generate ground truth data associations for training. 

We trained the small \senet\ and flow networks to predict a frame 0.27 seconds in the future to allow for large motions. We use the commanded  joint angles and velocities as the control ($n=14$). Due to the limited quantity of data, we make a modification to speed up training: we provide the \senet\ with the ground truth mask of the robot arm. While this makes the problem easier, the network still has to segment the object and jointly predict the arm and the object's motion. 

Fig.~\ref{fig:realpokesequence} shows a representative sequence from testing on a held out set of pokes - we cascade our predictions forward for more than one second. While the errors increase over time, our network is able to segment and predict consistent motion for the object and the arm. In comparison, the flow network performs very poorly (though it does have a harder problem as the ground truth arm label is not fed to it). Once again, we suggest the reader to look at the supplementary video for more results. Overall, we believe that this is a strong proof of concept showing: 1) we can easily generate training data needed for our networks and 2) \senets\ are able to learn scene dynamics from limited real world data. In the future, we plan to collect more data and train larger networks to handle complicated dynamics such as toppling and falling.
 
%%% Local Variables:
%%% mode: latex
%%% TeX-master: "paper"
%%% End:

\section{Discussion}
\label{sec:discussion}
Learning ``intuitive'' models of the physical world from raw data is a promising alternative to explicitly designed physics-based models.  This is due to the fact that models learned from data are tightly coupled to perception and thus well suited for closed-loop control. Toward this long term goal, we introduced \senets, a deep learning model that predicts changes in an environment based on applied actions, parameterized as a series of rigid transforms applied to 3D points in the environment. \senets\ selectively learn to focus on parts of the scene where motion occurs, segmenting the scene into objects and predicting $\SE3$ motions for each distinct object. We showed that this separation works well in practice and results in strong performance on four simulated and one real robot task with multiple rigid bodies in motion. \senets\ are able to generalize across different scenes and produce results that are very consistent with the observed rigid motion, as compared to traditional flow networks.

There are several promising directions for future work. First, while our experiments indicate that \senets\ can learn predictive models from real data, we are confident that a larger data collection effort would enable us to train networks that generalize across many types of objects and scenes. A key area for improvement is in learning data associations, which we currently provide as part of the training setup. A first step towards this would be to formulate a loss function based on Iterative Closest Point matching, which is able to align close-by depth data. Another exciting direction is to use the ``No Penalty'' version of \senets\ to model non-rigid motion, exploring the use of strong regularization and locality priors in the masks \cite{New15Dyn} to improve efficiency and generalization. Other areas for future work include learning multi-step recurrent \senets\ for sequential prediction, and using \senet\ models for closed-loop control.
%
%Another exciting direction is to learn specific object properties such as height, mass, and surface friction from experience, \emph{without} providing explicit training signals (similar to the masks learned by \senets). 
%
%
%Currently, \senets\ do not model occlusions or fill-in data. An intuitive solution to handle this can be to extend the current paradigm of disentangled motion/mask computation by adding an explicit component that learns to fill-in/remove data. 

%%% Local Variables:
%%% mode: latex
%%% TeX-master: "paper"
%%% End:

\vspace{-2mm}
\section*{Acknowledgments}
\vspace{-1mm}
This work was funded in part by the National Science Foundation under contract numbers NSF-NRI-1227234 and NSF-NRI-1637479.
\vspace{-2mm}

%% Use plainnat to work nicely with natbib. 
\vspace{-2mm}
{\small
\bibliographystyle{IEEEtran}
\bibliography{references}

\begin{thebibliography}{10}
\providecommand{\url}[1]{#1}
\csname url@rmstyle\endcsname
\providecommand{\newblock}{\relax}
\providecommand{\bibinfo}[2]{#2}
\providecommand\BIBentrySTDinterwordspacing{\spaceskip=0pt\relax}
\providecommand\BIBentryALTinterwordstretchfactor{4}
\providecommand\BIBentryALTinterwordspacing{\spaceskip=\fontdimen2\font plus
\BIBentryALTinterwordstretchfactor\fontdimen3\font minus
  \fontdimen4\font\relax}
\providecommand\BIBforeignlanguage[2]{{%
\expandafter\ifx\csname l@#1\endcsname\relax
\typeout{** WARNING: IEEEtran.bst: No hyphenation pattern has been}%
\typeout{** loaded for the language `#1'. Using the pattern for}%
\typeout{** the default language instead.}%
\else
\language=\csname l@#1\endcsname
\fi
#2}}

\bibitem{todorov2012}
E.~Todorov, T.~Erez, and Y.~Tassa, ``Mujoco: A physics engine for model-based
  control,'' in \emph{IROS}.\hskip 1em plus 0.5em minus 0.4em\relax IEEE, 2012,
  pp. 5026--5033.

\bibitem{geoffroy2014}
P.~Geoffroy, N.~Mansard, M.~Raison, S.~Achiche, and E.~Todorov, ``From inverse
  kinematics to optimal control,'' in \emph{Advances in Robot
  Kinematics}.\hskip 1em plus 0.5em minus 0.4em\relax Springer, 2014, pp.
  409--418.

\bibitem{zhouconvex}
J.~Zhou, R.~Paolini, J.~A. Bagnell, and M.~T. Mason, ``A convex polynomial
  force-motion model for planar sliding: Identification and
  application.''\hskip 1em plus 0.5em minus 0.4em\relax IEEE, 2016.

\bibitem{battaglia2013simulation}
P.~W. Battaglia, J.~B. Hamrick, and J.~B. Tenenbaum, ``Simulation as an engine
  of physical scene understanding,'' \emph{PNAS}, 2013.

\bibitem{sanborn2013reconciling}
A.~N. Sanborn, V.~K. Mansinghka, and T.~L. Griffiths, ``Reconciling intuitive
  physics and newtonian mechanics for colliding objects.'' \emph{Psychological
  review}, vol. 120, no.~2, p. 411, 2013.

\bibitem{baillargeon2004infants}
R.~Baillargeon, ``Infants' physical world,'' \emph{Current directions in
  psychological science}, vol.~13, no.~3, pp. 89--94, 2004.

\bibitem{whitney2016disentangled}
W.~Whitney, ``Disentangled representations in neural models,'' \emph{arXiv
  preprint arXiv:1602.02383}, 2016.

\bibitem{toussaint2009robot}
M.~Toussaint, ``Robot trajectory optimization using approximate inference,'' in
  \emph{ICML}.\hskip 1em plus 0.5em minus 0.4em\relax ACM, 2009, pp.
  1049--1056.

\bibitem{deisenroth2011pilco}
M.~Deisenroth and C.~E. Rasmussen, ``Pilco: A model-based and data-efficient
  approach to policy search,'' in \emph{ICML}, 2011, pp. 465--472.

\bibitem{kopicki2016learning}
M.~Kopicki, S.~Zurek, R.~Stolkin, T.~Moerwald, and J.~L. Wyatt, ``Learning
  modular and transferable forward models of the motions of push manipulated
  objects,'' \emph{Autonomous Robots}, pp. 1--22, 2016.

\bibitem{boots2014learning}
B.~Boots, A.~Byravan, and D.~Fox, ``Learning predictive models of a depth
  camera \& manipulator from raw execution traces,'' in \emph{ICRA}.\hskip 1em
  plus 0.5em minus 0.4em\relax IEEE, 2014, pp. 4021--4028.

\bibitem{Boh16Int}
J.~Bohg, K.~Hausman, B.~Sankaran, O.~Brock, D.~Kragic, S.~Schaal, and
  G.~Sukhatme, ``Interactive perception: Leveraging action in perception and
  perception in action,'' \emph{arXiv preprint arXiv:1604.03670}, 2016.

\bibitem{watter2015embed}
M.~Watter, J.~Springenberg, J.~Boedecker, and M.~Riedmiller, ``Embed to
  control: A locally linear latent dynamics model for control from raw
  images,'' in \emph{NIPS}, 2015, pp. 2728--2736.

\bibitem{wahlstrom2015pixels}
N.~Wahlstr{\"o}m, T.~B. Sch{\"o}n, and M.~P. Deisenroth, ``From pixels to
  torques: Policy learning with deep dynamical models,'' \emph{arXiv preprint
  arXiv:1502.02251}, 2015.

\bibitem{lerer2016learning}
A.~Lerer, S.~Gross, and R.~Fergus, ``Learning physical intuition of block
  towers by example,'' \emph{arXiv preprint arXiv:1603.01312}, 2016.

\bibitem{fragkiadaki2015learning}
K.~Fragkiadaki, P.~Agrawal, S.~Levine, and J.~Malik, ``Learning visual
  predictive models of physics for playing billiards,'' \emph{arXiv preprint
  arXiv:1511.07404}, 2015.

\bibitem{mottaghi2016newtonian}
R.~Mottaghi, H.~Bagherinezhad, M.~Rastegari, and A.~Farhadi, ``Newtonian scene
  understanding: Unfolding the dynamics of objects in static images,'' in
  \emph{CVPR}, 2016, pp. 3521--3529.

\bibitem{agrawal2015learning}
P.~Agrawal, J.~Carreira, and J.~Malik, ``Learning to see by moving,'' in
  \emph{ICCV}, 2015, pp. 37--45.

\bibitem{finn2016unsupervised}
C.~Finn, I.~Goodfellow, and S.~Levine, ``Unsupervised learning for physical
  interaction through video prediction,'' in \emph{NIPS}, 2016.

\bibitem{kulkarni2015deep}
T.~D. Kulkarni, W.~F. Whitney, P.~Kohli, and J.~Tenenbaum, ``Deep convolutional
  inverse graphics network,'' in \emph{NIPS}, 2015, pp. 2530--2538.

\bibitem{yang2015weakly}
J.~Yang, S.~E. Reed, M.-H. Yang, and H.~Lee, ``Weakly-supervised disentangling
  with recurrent transformations for 3d view synthesis,'' in \emph{NIPS}, 2015,
  pp. 1099--1107.

\bibitem{handa2016gvnn}
A.~Handa, M.~Bloesch, V.~P{\u{a}}tr{\u{a}}ucean, S.~Stent, J.~McCormac, and
  A.~Davison, ``gvnn: Neural network library for geometric computer vision,''
  in \emph{Computer Vision--ECCV 2016 Workshops}, 2016, pp. 67--82.

\bibitem{gregor2015draw}
K.~Gregor, I.~Danihelka, A.~Graves, and D.~Wierstra, ``Draw: A recurrent neural
  network for image generation,'' \emph{arXiv:1502.04623}, 2015.

\bibitem{ba2014multiple}
J.~Ba, V.~Mnih, and K.~Kavukcuoglu, ``Multiple object recognition with visual
  attention,'' \emph{arXiv preprint arXiv:1412.7755}, 2014.

\bibitem{jaderberg2015spatial}
M.~Jaderberg, K.~Simonyan, A.~Zisserman, \emph{et~al.}, ``Spatial transformer
  networks,'' in \emph{NIPS}, 2015, pp. 2008--2016.

\bibitem{long2015fully}
J.~Long, E.~Shelhamer, and T.~Darrell, ``Fully convolutional networks for
  semantic segmentation,'' in \emph{CVPR}, 2015, pp. 3431--3440.

\bibitem{hinterstoisser2012accv}
S.~Hinterstoisser, V.~Lepetit, S.~Ilic, S.~Holzer, G.~Bradski, K.~Konolige, ,
  and N.~Navab, ``Model based training, detection and pose estimation of
  texture-less 3d objects in heavily cluttered scenes,'' in \emph{ACCV}, 2012.

\bibitem{collobert2011torch7}
R.~Collobert, K.~Kavukcuoglu, and C.~Farabet, ``Torch7: A matlab-like
  environment for machine learning,'' in \emph{BigLearn, NIPS Workshop}, 2011.

\bibitem{kingma2014adam}
D.~Kingma and J.~Ba, ``Adam: A method for stochastic optimization,''
  \emph{arXiv preprint arXiv:1412.6980}, 2014.

\bibitem{ioffe2015batch}
S.~Ioffe and C.~Szegedy, ``Batch normalization: Accelerating deep network
  training by reducing internal covariate shift,'' \emph{arXiv preprint
  arXiv:1502.03167}, 2015.

\bibitem{agrawal2016learning}
P.~Agrawal, A.~Nair, P.~Abbeel, J.~Malik, and S.~Levine, ``Learning to poke by
  poking: Experiential learning of intuitive physics,'' \emph{arXiv preprint
  arXiv:1606.07419}, 2016.

\bibitem{schmidt2014dart}
T.~Schmidt, R.~A. Newcombe, and D.~Fox, ``Dart: Dense articulated real-time
  tracking.'' in \emph{RSS}, 2014.

\bibitem{New15Dyn}
R.~Newcombe, D.~Fox, and S.~Seitz, ``{DynamicFusion}: Reconstruction and
  tracking of non-rigid scenes in real-time,'' in \emph{CVPR}, 2015.

\end{thebibliography}
}

\end{document}